\def\year{2022}\relax
\documentclass[letterpaper]{article} 
\usepackage{aaai22}  
\usepackage{times}  
\usepackage{helvet}  
\usepackage{courier}  
\usepackage[hyphens]{url}  
\usepackage{graphicx} 
\usepackage{natbib}  
\usepackage{caption} 
\usepackage[switch]{lineno}

\DeclareCaptionStyle{ruled}{labelfont=normalfont,labelsep=colon,strut=off} 
\usepackage{algorithm}
\usepackage{algorithmic}
\usepackage{amssymb}
\usepackage[inline]{enumitem}
\usepackage{relsize}
\usepackage{stmaryrd}
\usepackage{arydshln}
\usepackage[normalem]{ulem}
\usepackage{microtype}
\usepackage{amsmath}
\usepackage{amsfonts}
\usepackage{booktabs}
\usepackage{multirow}
\usepackage{color,xcolor}
\usepackage{soul}
\usepackage{tabularx}
\newcommand{\dashrule}[1][black]{%
  \color{#1}\rule[\dimexpr.5ex-.2pt]{4pt}{.4pt}\xleaders\hbox{\rule{4pt}{0pt}\rule[\dimexpr.5ex-.2pt]{4pt}{.4pt}}\hfill\kern0pt%
}
\definecolor{casered}{rgb}{0.7,0,0}
\definecolor{caseblue}{rgb}{0.27451,0.5098,0.70588}
\usepackage{newfloat}
\usepackage{listings}
\floatstyle{ruled}
\newfloat{listing}{tb}{lst}{}
\floatname{listing}{Listing}
\title{Knowledge Bridging for Empathetic Dialogue Generation}
\author{
    Qintong Li\textsuperscript{\rm 1,3},  
    Piji Li\textsuperscript{\rm 2}\correspondauthor, 
    Zhaochun Ren\textsuperscript{\rm 1}\correspondauthor, 
    Pengjie Ren\textsuperscript{\rm 1}, 
    Zhumin Chen\textsuperscript{\rm 1}\\
}
\affiliations{
    \textsuperscript{\rm 1}School of Computer Science and Technology, Shandong University, Qingdao, China\\
    \textsuperscript{\rm 2}Tencent AI Lab, Shenzhen, China\\
    \textsuperscript{\rm 3}Department of Computer Science, The University of Hong Kong, Hong Kong SAR, China\\
    qtleo@outlook.com, 
    \{zhaochun.ren, chenzhumin\}@sdu.edu.cn, 
    lipiji.pz@gmail.com, 
    jay.ren@outlook.com
}

\usepackage{bibentry}

\begin{document}
\maketitle

\begin{abstract}
    Lack of external knowledge makes empathetic dialogue systems difficult to perceive implicit emotions and learn emotional interactions from limited dialogue history.
    To address the above problems, we propose to leverage external knowledge, including commonsense knowledge and emotional lexical knowledge, to explicitly understand and express emotions in empathetic dialogue generation.
    We first enrich the dialogue history by jointly interacting with external knowledge and construct an emotional context graph. 
    Then we learn emotional context representations from the knowledge-enriched emotional context graph and distill emotional signals, which are the prerequisites to predicate emotions expressed in responses.
    Finally, to generate the empathetic response, we propose an emotional cross-attention mechanism to learn the emotional dependencies from the emotional context graph.
    Extensive experiments conducted on a benchmark dataset verify the effectiveness of the proposed method. 
    In addition, we find the performance of our method can be further improved by integrating with a pre-trained model that works orthogonally.
\end{abstract}

\section{Introduction}
Studies on social psychology suggest that \textit{empathy} is a crucial factor towards a more humanized dialogue system~\cite{zech2005talking}.
Although plenty of researchers have attempted to control the emotional content of response either through an explicitly assigned emotional label~\cite{zhou2018mojitalk,zhou2018emotional,Wang018,song2019generating,ShenF20} or through a general term to encourage higher levels of affect~\cite{AsgharPHJM18}, it is still challenging for chatbots to conduct empathetic dialogues without the explicit emotion labels~(empathetic dialogue problem)~\cite{zhou2018emotional,rashkin2019towards}.
Several recent works have been proposed to address the empathetic dialogue problem based on multi-task learning~\cite{Rashkin18,rashkin2019towards,wei2019emotion,lin2020caire}, the mixture of experts~\cite{LinMSXF19}, emotion mimicry~\cite{MajumderHPLGGMP20}, or multi-resolution user feedback~\cite{LiCRRTC20}.

\begin{figure}[!t]
 \centering
 \includegraphics[width=0.41\textwidth]{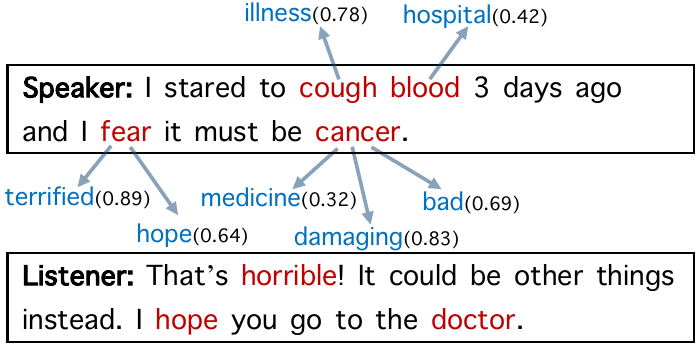}
 \caption{An example of empathetic dialogues with external knowledge from \textsc{EmpatheticDialogues}~\cite{rashkin2019towards}. Emotion-related words in the dialogue are highlighted in \textcolor{casered}{red} color, whereas emotion-related concepts are marked in \textcolor{caseblue}{blue}. Numbers in parentheses denote emotional intensity values. 
}
 \label{fig:example}
\end{figure}

However, an unheeded deep concern is that humans usually rely on experience and external knowledge to acknowledge and express implicit emotions~\cite{zhong2019knowledge}.
Figure~\ref{fig:example} shows a real-world example of empathetic dialogues. 
If we use non-stopwords of speaker's input as queries to acquire knowledge via external knowledge, we can obtain various emotion-related concepts along with their emotional intensity values, which play a crucial role in emotion understanding for empathetic dialogue systems.

\begin{figure}[!t]
 \centering
 \includegraphics[width=0.45\textwidth]{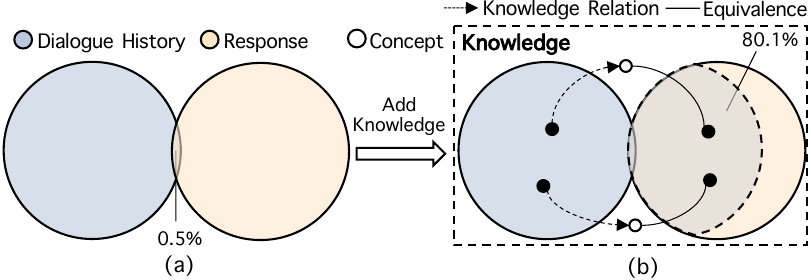}
 \caption{Relationships among the dialogue history, responses, and external knowledge.}
 \label{fig:knowledge_ratio}
\end{figure}

To exploit this phenomenon more concretely, we quantitatively investigate effects of external knowledge in understanding emotions on an empathetic dialogue corpus, i.e., \textsc{EmpatheticDialogues}~\cite{rashkin2019towards}.
Figure~\ref{fig:knowledge_ratio}(a) depicts that the response has almost NO non-stopword overlapping~(0.5\% of dialogue samples) with the dialogue history. 
This phenomenon implies that humans need to infer more external knowledge to conduct empathetic dialogues.
By contrast, if we incorporate external knowledge (i.e., emotion-related concepts) into the system, we observe that for most dialogue samples (80.1\%) chatbots can directly obtain hints from the knowledge paths started by the non-stop tokens of the dialogue history (shown in Figure~\ref{fig:knowledge_ratio}(b)).
Hence, external knowledge is essential in acquiring useful emotional knowledge and improving the performance of empathetic dialogue generation. 
However, emotion perception and representation from external knowledge is still problematic for empathetic dialogue generation.

\begin{figure}[!t]
  \centering
  \includegraphics[width=0.43\textwidth,height=3.9cm]{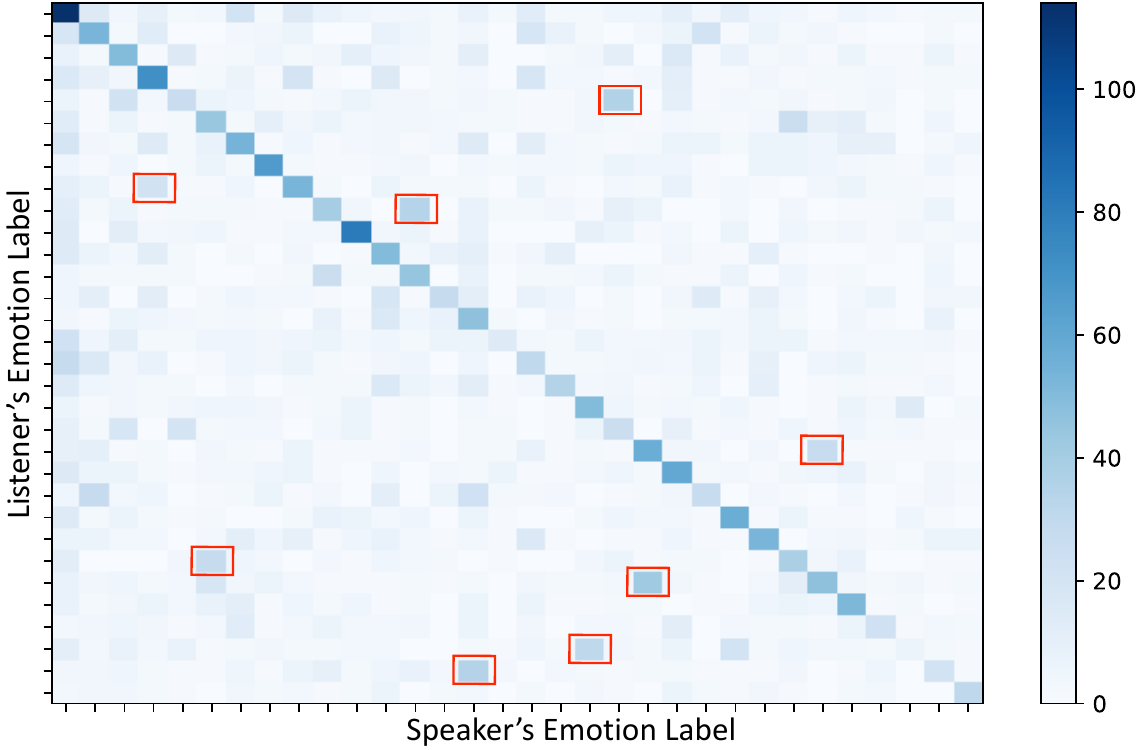}
  \caption{Emotion transition patterns.} 
  \label{fig:emotion_transition}
 \end{figure}

During the investigations, we observe another phenomenon that emotional dependency and emotional inertia commonly appear with external knowledge in empathetic conversations.
We label utterances with a CNN-based emotion classifier~\cite{Kim14}, and visualize the emotion transitions from speakers to the listeners in Figure~\ref{fig:emotion_transition}. 
In Figure~\ref{fig:emotion_transition}, the darker diagonal grids show that listeners tend to mirror the emotion of their interlocutors to build rapport~\cite{Navarretta16}. 
Moreover, there are also some complex emotional transition patterns besides the diagonal direction (in red frame).
Therefore, intuitively, it is crucial to model emotional dependencies between interlocutors.

To this end, we propose a \textbf{K}nowledge-aware \textbf{EMP}athetic dialogue generation method~(\textbf{KEMP}).
It consists of three components: an emotional context graph, an emotional context encoder, and an emotion-dependency decoder. 
The emotional context graph is constructed via integrating the dialogue history with external knowledge. 
The emotional context encoder employs the graph-aware transformer to learn the graph embeddings, and propose an emotional signal perception procedure to perceive context emotions that lead the response generation.
Conditioned on the knowledge-enriched context graph, the emotion-dependency decoder particularly models emotion dependencies to generate empathetic response.
A multi-task learning framework is applied to jointly optimize our objectives.

Conducted on the benchmark dataset \textsc{EmpatheticDialogues}~\cite{rashkin2019towards}, extensive experimental results demonstrate the effectiveness of KEMP in terms of both automatic and human evaluations.

In summary, our contributions are as follows:
\begin{enumerate*}[label=(\alph*)]
    \item We propose KEMP which is able to accurately perceive and appropriately express implicit emotions. To the best of our knowledge, this is the first attempt to leverage external knowledge to enhance empathetic dialogue generation.
    \item We design an emotional context encoder and an emotion-dependency decoder to learn the emotional dependencies between the emotion-enhanced representations of the dialogue history and target response.
    \item We conduct extensive experiments and analyses to demonstrate the effectiveness of KEMP.\footnote{Code and dataset are available at \url{http://github.com/qtli/KEMP}.}
\end{enumerate*}

\section{Related work}
\subsection{Emotional dialogue generation}
With the rise of data-driven learning approaches~\cite{SutskeverVL14, vaswani2017attention}, open-domain dialogue generation models have seen growing interests in recent years~\cite{VinyalsL15, ShangLL15, SerbanLCP16, LiMRJGG16, ZhouYHZXZ18, DinanRSFAW19}. 
To control the emotional content of the target output, recent approaches generate emotional responses conditioning on a manually specified label~\cite{zhou2018emotional,LiS18, zhou2018mojitalk, huang2018automatic,wei2019emotion,colombo2019affect, ShenF20}.
However, existing emotional dialogue models purely focus on whether the generated response matches a predetermined emotion, whereas in real-world scenarios the listener is capable to infer the emotion of the speaker~\cite{rashkin2019towards}.

\subsection{Empathetic dialogue generation}
Unlike the task of emotional dialogue generation, the task of empathetic dialogue generation avoids an additional step of determining which emotion type to respond explicitly~\cite{skowron2013affect}.
Several works~\cite{Rashkin18, zhong2019affect, Shin19, ChatterjeeGCSGA19, rashkin2019towards, Santhanam19, LinMSXF19, lin2020caire, zhong2020, MajumderHPLGGMP20, LiCRRTC20} have attempted to make dialogue models more empathetic.
\citeauthor{rashkin2019towards}~\shortcite{rashkin2019towards} combine existing models in different ways to produce empathetic responses.
\citeauthor{LinMSXF19}~\shortcite{LinMSXF19} softly combine the possible emotional responses from several separate experts. 
\citeauthor{MajumderHPLGGMP20}~\shortcite{MajumderHPLGGMP20} considere of this polarity-based emotion clusters and emotional mimicry.
\citeauthor{LiCRRTC20}~\shortcite{LiCRRTC20} propose a multi-resolution adversarial framework which considers multi-granularity emotion factors and users' feedback.

Besides the advancements in empathetic dialogue models, the emergence of new emotion-labelled dialogue corpora have also contributed to this research field~\cite{li2017dailydialog,HsuCKHK18,rashkin2019towards}.
\citeauthor{rashkin2019towards}~\shortcite{rashkin2019towards} consider a richer and evenly distributed set of emotions and release a dataset \textsc{EmpatheticDialogues}, where a listener responds to a speaker who is under an emotional situation in an empathetic way.
In this work, we investigate how to leverage external knowledge to explicitly improve the emotional understanding and expression in the task of empathetic dialogue generation on the dataset of \textsc{EmpatheticDialogues}.

\section{Preliminaries}
In this work, external knowledge serves as the bridge to improve emotion perception and emotion expression capabilities.
Therefore, we first introduce the two-type knowledge sources used in KEMP: the commonsense knowledge ConceptNet~\citep{speer2017conceptnet} and the emotional lexicon NRC\_VAD~\citep{mohammad2018obtaining}.

\textbf{ConceptNet} is a large-scale knowledge graph that describes general human knowledge in natural language, playing an effective role in sentiment-related task~\citep{GhosalHRMMP20}.
It comprises 5.9M tuples, 3.1M concepts, and 38 relations.
We denote each tuple (head concept, relation, tail concept, confidence score) as \(\tau=(x,r,c,s)\), e.g., $\left \langle \mathrm{birthday}, \mathrm{RelatedTo}, \mathrm{happy}, 0.19 \right \rangle$.

\textbf{NRC\_VAD} is a lexicon of VAD (Valence-Arousal-Dominance) vectors with $3$-dimensions ($V_a$, $A_r$, $D_o$) for 20k English words, e.g., the VAD vector of word ``nice'' is: $[0.93, 0.442, 0.65]$.
VAD vectors are culture-independent and widely adopted in Psychology~\citep{mehrabian1996pleasure}.
The interpretations of VAD vectors are presented in Table~\ref{tab:VAD}.

\begin{table}[h]
\caption{\label{tab:VAD}Interpretations of VAD vectors.}
\centering
\begin{tabular}{l|c|l}
\toprule
\textbf{Dimensions} & \textbf{Values} & \textbf{Interpretations}\\
\midrule
Valence & $[0, 1]$ & Negative - Positive \\
Arousal & $[0, 1]$ & Calm - Excited \\
Dominance & $[0, 1]$ & Submissive - Dominant \\
\bottomrule
\end{tabular}
\end{table}
To highlight emotional information, we adopt NRC\_VAD to compute emotion intensity values~\citep{zhong2019knowledge} for dialogue words and external concepts $x$: 
\begin{equation} 
    \eta(x) = \textrm{\textit{min-max}}(\left \| V_a(x)-\frac{1}{2}, \frac{A_r(x)}{2} \right \|_2)\text{,} \label{emotion_intensity} 
\end{equation}
where $\textit{min-max()}$ is min-max normalization; $\left \| . \right \|_k$ denotes $L_k$ norm; $V_a(x)$ and $A_r(x)$ denote the values of valence and arousal dimensions in VAD vector of word $x$,  respectively.   
If $x$ is not in NRC\_VAD, $\eta(x)$ will be set to 0.

We inject concepts with higher emotion intensity values from ConceptNet into KEMP to help emotion perception and expression.

\section{Method}
\subsection{Overview}
\label{sec41}

\begin{figure*}[!t]
    \centering
    \includegraphics[width=0.9\textwidth]{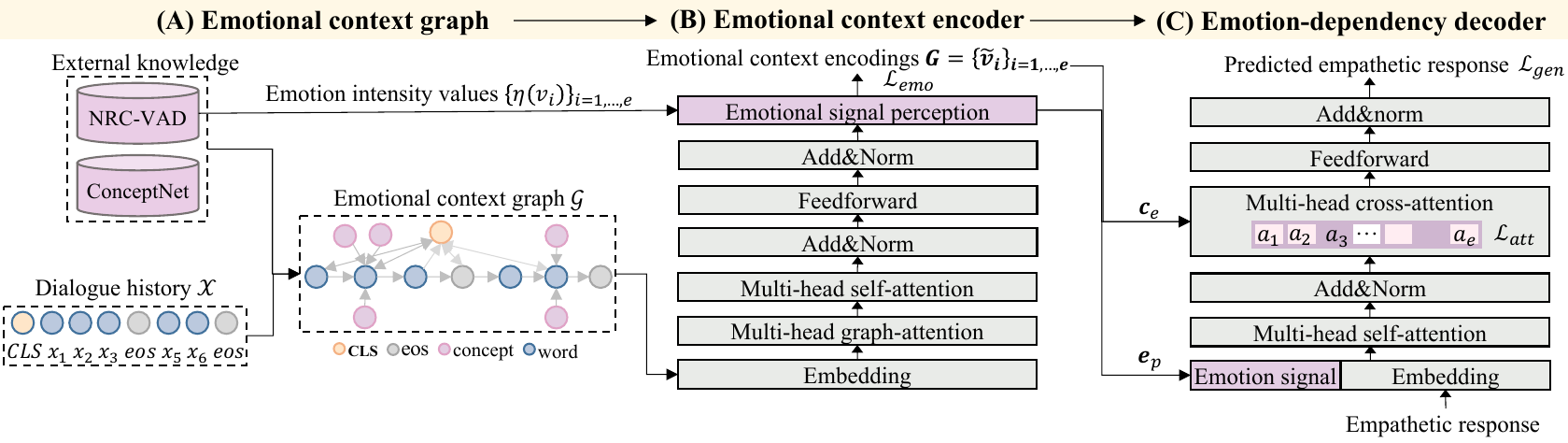}
    \caption{An overall architecture of KEMP. Model inputs are in the dotted box.}
    \label{fig:model}
\end{figure*}

We provide a general overview of KEMP in Figure~\ref{fig:model}. KEMP consists of $3$ phases: (A)~{emtional context graph}, (B)~{emotional context encoder}, and (C)~{emotion-dependency decoder}.
To summarize, we are given a dialogue history with $M$ utterances, i.e., $\mathcal{D}=[X_1, \dots, X_M]$, as the input, where the $i$-th utterance $X_i=[x_0^i,\ldots,x_{m_i}^i]$ is a sequence of $m_i$ words.
In phrase (A), we enrich the dialogue history $\mathcal{D}$ with external knowledge into an emotional context graph $\mathcal{G}$.
In phrase (B), emotional signals $e_p$ of $\mathcal{D}$ are distilled based on the embeddings and emotion intensity values from $\mathcal{G}$.
Given $e_p$ and $\mathcal{G}$, phrase (C) incorporates an emotional cross-attention mechanism to selectively learn the emotional dependencies. Subsequently, we generate an empathetic response $\mathcal{Y} = [y_1, \dots, y_n]$ with appropriate emotion and informative content.

\subsection{Emotional context graph}
\label{sec_graph}

We construct emotional context graph $\mathcal{G}$ by interacting with two-type external knowledge sources.
Following~\citet{LiCRRTC20}, we flat dialogue history into a long word sequence and insert a \texttt{CLS} token at the start of the token sentence, i.e., $\mathcal{X}=[\texttt{CLS},x_1,\ldots,x_m]$. 
For each non-stopword word $x_i \in \mathcal{X}$, we first retrieve a set of candidate tuples ${\rm T}_i =\left \{ \tau_i^k=(x_i,r_i^k,c_i^k, s_i^k)\right \}_{k=1,\ldots,K}$ from ConceptNet. 
Then we adopt three heuristic steps to refine the emotion-related knowledge:
(1)\if0\textbf{relation selection.}\fi We extract a subset $\hat {\rm T}_i \subset {\rm T}_i$ by filtering tuples with relevant relations for empathetic response (e.g., ``Causes'') and adequate confidence score (i.e., $s_i^k>0.1$).
(2)\if0\textbf{concept selection.}\fi
 We rank tuples by the emotion intensity values $\{\eta(c_i^k)\}_{k=1,\ldots,K}$ of retrieved concepts $\{c_i^k\}_{k=1,\ldots,K}$.
For each word $x_i$, we select top $K'$ tuples as the emotional knowledge subgraph. 
(3)\if0\textbf{edge connection.}\fi We apply 3 types of directed edges to connect vertices: \begin{enumerate*}[label=(\roman*)]
    \item \textit{temporary} edges between two successive words;
    \item \textit{emotion} edges between a word $x_i$ and its emotional concepts $c_i^k$;
    \item \textit{globality} edges between \texttt{CLS} token and other vertices.
\end{enumerate*}

Finally, the dialogue history is enriched by emotional knowledge and represented as the emotional context graph $\mathcal{G}$.
The words $x \in \mathcal{X}$ and the emotional concepts constitute the vertices $V=\{v_i\}_{i=1,\ldots,e}$ of $\mathcal{G}$, where $e$ is the number of vertices.
The above edges among vertices are set to 1 in the adjacency matrix $\mathcal{A}$ of $\mathcal{G}$.

\subsection{Emotional context encoder} 
\label{encoder}

\paragraph{Emotional context graph encoding.} We first use a word embedding layer and a positional embedding layer~\citep{vaswani2017attention} to convert each vertice $v_i \in \mathcal{G}$ into vectors $\textbf{E}_w(v_i) \in \mathbb{R}^d$ and $\textbf{E}_p(v_i) \in \mathbb{R}^d$, where $d$ is the dimensionality of embeddings.
In the multi-turn dialogue settings, distinguishing vertices in dialogue history or external knowledge is helpful. So we incorporate the vertice state embedding $\textbf{E}_v(v_i)$ for vertice $v_i$. 
The vector representation of vertices $v_i$ is the composition of three types of embeddings:

\begin{equation}
    \mathbf{v}_i = \textbf{E}_w(v_i) + \textbf{E}_p(v_i) + \textbf{E}_v(v_i).
\end{equation}

Then we apply a multi-head graph-attention mechanism to update the vertice representations with emotional knowledge.
Specifically, each vertice ${\bf v}_i$ is contextualized by attending to all its immediate neighbours $\{ {\bf v}_j \}_{j\in\mathcal{A}_i}$:

\begin{equation}
\begin{aligned}
    \mathbf{\hat{v}}_i &= {\bf v}_i + \bigparallel_{n=1}^H \sum_{j \in \mathcal{A}_i}\alpha^n_{ij} \mathbf{W}^n_{v} \mathbf{v}_j,  \\
    \alpha^n_{ij} &= a^n({\bf v}_i,{\bf v}_j),
\end{aligned}
\end{equation}

\noindent where $\|$ denotes the concatenation of $H$ attention heads, $\mathcal{A}_i$ denotes the neighborhood of $v_i$ in the adjacency matrix $\mathcal{A}$, and $a^n$ represents the self-attention mechanism of the $n$-th head in the following format:
\begin{equation}
    a^n({\bf q}_i,{\bf k}_j) = \frac{\exp(({\mathbf{W}^n_{q}\mathbf{q}_i)^{\top}\mathbf{W}^n_{k}{\bf k}_j})}{\sum_{z \in \mathcal{A}_i}\exp(({\mathbf{W}^n_{q}{\bf q}_i)^{\top}\mathbf{W}^n_{k}\mathbf{k}_z})},
\end{equation}
where $\mathbf{W}^n_{q}\in \mathbb{R}^{d_h \times d_h}$, $\mathbf{W}^n_{k}\in \mathbb{R}^{d_h \times d_h}$ are the linear transformations. $d_h=d/H$ is the dimension of each head.

As previous operations are only conducted to the local context (i.e., immediate neighbours), we update the vertex representations with the global context information (i.e., all other vertices) to model global interactions.
Concretely, we use transformer layers~\citep{vaswani2017attention} to inject global information for all vertices $\{\mathbf{\hat{v}}_i\}_{i=1,\ldots,m}$:

\begin{small}
\begin{eqnarray}
    \mathbf{h}_i^l &=& \textrm{LayerNorm}(\mathbf{\hat{v}}_i^{l-1} + \textrm{MHAtt}(\mathbf{\hat{v}}_i^{l-1})), \\
    \mathbf{\hat{v}}_i^l &=& \textrm{LayerNorm}(\mathbf{h}_i^l + \textrm{FFN}(\mathbf{h}_i^l)),
\end{eqnarray} 
\end{small}

\noindent where \textrm{LayerNorm} is the Layer Normalization trick~\citep{BaKH16}; \textrm{MHAtt} is the multi-head self-attention sub-layer consiting of $H$ attention heads; \textrm{FFN} is a two-layer feed-forward network with ReLU as hidden activation function.
The emotional context graph $\mathcal{G}$ is represented as $\mathbf{G} = \{ \mathbf{\tilde{v}}_i \}_{i=1,\ldots,e}$, where $\mathbf{\tilde{v}}_i=\mathbf{\hat{v}}_i^l$. 

\paragraph{Emotional signal perception.} Our model learns the emotional signals from the emotional context graph to guide the empathetic response generation. 
The emotional signal representation $\mathbf{c}_e  \in \mathbb{R}^{d}$ is the weighted summation of vertice representations $\{ \mathbf{\tilde{v}}_i \}_{i=1,\ldots,e}$ on their emotion intensity values $\{\eta(v_i)\}_{i=1,\ldots,e}$:
\begin{equation}
    \mathbf{c}_e = \sum_{i=1}^m\frac{\exp(\eta_i)}{\sum_{j=1}^e\exp(\eta_j)}\mathbf{\tilde{v}}_i .
 \end{equation}

Then a linear layer with $\operatorname{softmax}$ operation projects the vector $\mathbf{c}_e$ into an emotion category distribution $P_e$ over the emotion label to identify the emotional signal for the empathetic response:

\begin{eqnarray}
    \mathbf{e}_p = &\mathbf{W}_{e}\mathbf{c}_e,\\
    P_e(e|\mathcal{G}) = &\textrm{\text{softmax}}(\mathbf{e}_p),
\end{eqnarray}

\noindent where $\mathbf{W}_{e} \in \mathbb{R}^{q \times d}$ and $q$ is the number of emotion categories.
During training, we employ negative log-likelihood as the emotion perception loss to conduct the parameter learning:

\begin{equation}
    \mathcal{L}_{emo} = -\log(P_e(e=e^*|\mathcal{G})),
\label{eq:mle_loss}
\end{equation}

\noindent where $e^*$ denotes the ground truth emotion label of dialogue history and $e$ denotes the predicted label. Together with the emotional context encodings $\mathbf{G}$, emotional vectors $\mathbf{e}_p$ and $\mathbf{c}_e$ will be fed into the decoder as a crucial emotional signal to guild the empathetic response generation.

\subsection{Emotion-dependency decoder} 
\label{sec:efrg}

Starting from the intermediate emotional signal $\mathbf{e}_p \in \mathbb{R}^{1 \times q}$, we propose an emotion-dependency decoder to generate the target word sequentially. 
To acquire emotion dependencies from $\mathcal{G}$ and control empathetic response expression, we linearly transform $\mathbf{e}_p$ to $\mathbf{e}'_p$ via $\mathbf{e}'_p = \mathbf{W}_z \mathbf{e}_p + \mathbf{b}_z$. 
At the $j$-th decoding step, $\mathbf{e}'_p$ is concatenated with the embeddings of words $[y_1,\ldots,y_{j-1}]$ into $[\mathbf{y}_0,\ldots,\mathbf{y}_{j-1}]$, where $\mathbf{y}_0=\mathbf{e}'_p$.
We then feed the embeddings into the response decoder.

Our decoder is built based on Transformer layers. 
Specially, to improve the emotional dependencies between the emotional context graph and target empathetic response, we design two emotional strategies, i.e., \textit{incorporating emotional features} and \textit{enforcing emotional attention loss} at the cross-attention sub-layer.

\paragraph{Incorporating emotional features.} To capture dialogue context vector $\mathbf{g}_s$ from emotional context graph $\mathcal{G}$, we compute the attention score between the last prediction word $\mathbf{y}_j$ and vertices $\{\mathbf{\tilde{v}}_i\}_{i=1,\ldots,e}$ as follows:

\begin{align} 
    a^n({\bf y}_{j-1} &,\mathbf{\tilde{v}}_i) = \frac{\exp(({\mathbf{W}^n_{c}\mathbf{\tilde{v}}_i)^{\top}\mathbf{W}^n_{r}{\bf y}_{j-1}})}{\sum_{v_z \in \mathcal{G}}\exp(({\mathbf{W}^n_c\mathbf{\tilde{v}}_z)^{\top}\mathbf{W}^n_{r}\mathbf{y}_{j-1}})},\\
    \mathbf{g}_s &= \bigparallel_{n=1}^Ha^n({\bf y}_{j-1},\mathbf{\tilde{v}}_i)\mathbf{W}^n_u\mathbf{\tilde{v}}_i,
\end{align}

\noindent where $H$ is the number of attention heads.
To improve the empathy expression of response, 
we concatenate the context vector $\mathbf{g}_s$ with the emotional signals $\textbf{c}_e$ into an emotional context vector $\mathbf{c}$, i.e., $\mathbf{c} = [\mathbf{g}_s;\mathbf{c}_e]$.

Then we feed the last word representation $\mathbf{y}_{j-1}$ and vector $\mathbf{c}$ to a two-layer feed-forward network, which has a ReLU activation function and a highway layer normalization, so we have:

\begin{align}
    \mathbf{s}_{j-1} &= \text{LayerNorm}(\mathbf{y}_{j-1} + \mathbf{c}), \\
    \mathbf{y}_{j} &=  \text{LayerNorm}(\mathbf{s}_{j-1} + \text{FFN}(\mathbf{s}_{j-1})),
\end{align}

\paragraph{Enforcing emotional attention loss.} Since humans naturally pay extra attention to the emotional salient information during a conversation~\citep{LiCRRTC20}, we enforce an emotional attention loss to focus on those vertices with higher emotion intensity values:

\begin{align}
        a_{i} &= \sum_n^H a^n(\mathbf{y}_{j-1},\mathbf{v}_i)/H, \\
     \mathcal{L}_{att} &= \frac{1}{e}\sum_{i=1}^e(\eta(v_i)-a_i)^2,
       \label{eq:att_loss}
\end{align}

\noindent Then the generator yields the distribution over the vocabulary $\mathcal{V}$ for the $j$-th word:
\begin{equation} 
    P_\mathcal{V}(y_j \mid \mathbf{y}_{0:j-1},\mathbf{G}) = \text{softmax}(\mathbf{W}_v\mathbf{y}_j + \mathbf{b}_v),
\end{equation}
where $\mathbf{W}_v \in \mathbb{R}^{|\mathcal{V}|\times d}$, $\mathbf{b}_v \in \mathbb{R}^{|\mathcal{V}|}$ are trainable parameters.

By using external concepts, we compute a probability $p_g$ of copying from vertices $\{v_i\}_{i=1,\ldots,e}$ in the graph $\mathcal{G}$ in a manner similar to~\citet{SeeLM17} and derive the final probability distribution $P(y_j)$:

\begin{align} 
    p_{gen} &= \sigma(\mathbf{W_g}\mathbf{y}_{j} + b_g), \\
    P(y_j) &= p_gP_\mathcal{V}(y_j) + (1-p_g)\sum_{i:v_i=y_j}a_i,
\end{align}

\noindent where $\mathbf{W_g} \in \mathbb{R}^{d}$ and $b_g \in \mathbb{R}$ are trainable parameters; $\sigma(\cdot)$ is the sigmoid activation function.
We use the negative log-likelihood of the ground-truth words $y^*_j$ as the generation loss function:

\begin{align}
  \mathcal{L}_{gen} &= -\sum_{j=1}^{n} \log P(y_j=y^{*}_{j} \mid y^*_{1,\ldots,j-1},\mathcal{G}).
  \label{eq:gen_loss}
\end{align}

Eventually, we adopt a multi-task learning framework to jointly minimize the emotion perception loss (Eq.~\ref{eq:mle_loss}), the emotional attention loss (Eq.~\ref{eq:att_loss}), and the generation loss (Eq.~\ref{eq:gen_loss}) as follows:
\begin{equation} 
     \mathcal{L} = \gamma_1\mathcal{L}_{emo} + \gamma_2\mathcal{L}_{gen} + \gamma_3\mathcal{L}_{att}.
\end{equation} 
where $\gamma_1,\gamma_2, \gamma_3$ are hyper-parameters.

\section{Experimental Settings}

\begin{table*}[!t]
\caption{Performance of all models.} 
\centering
\resizebox{2\columnwidth}{!}{
\begin{tabular}{l|c|c|c|c|c|c|c}
\toprule
\multirow{1}{*}{\textbf{Models}} &
\multirow{1}{*}{\textbf{Accuracy}} &
\multirow{1}{*}{\textbf{Perplexity}} &
\multirow{1}{*}{\textbf{Distinct-1}} &
\multirow{1}{*}{\textbf{Distinct-2}} &
\multirow{1}{*}{\textbf{Empathy}} &
\multirow{1}{*}{\textbf{Relevance}} &
\multirow{1}{*}{\textbf{Fluency}} 
\\
\midrule
Transformer~\citep{vaswani2017attention} & - & 37.73  & 0.47  & 2.04 & 3.11 & 3.47  & 3.66 \\
EmoPrepend-1~\citep{rashkin2019towards} & 33.28 & 38.30 & 0.46  & 2.08 & 3.23 & 3.51 & 3.67 \\
MoEL~\citep{LinMSXF19} & 32.00 & 38.04 & 0.44 & 2.10 & 3.37 & 3.78  & 3.64 \\
MIME~\citep{MajumderHPLGGMP20} & 34.24 & 37.09 & 0.47 & 1.91 & 3.38 & 3.66  & 3.63 \\
EmpDG~\citep{LiCRRTC20} & 34.31 & 37.29  & 0.46 & 2.02 & 3.45 & 3.88  & \textbf{3.67} \\
\midrule
KEMP &  \textbf{39.31} & \textbf{36.89} & \textbf{0.55} & \textbf{2.29}  & \textbf{3.49} & \textbf{3.92} & 3.65 \\
\bottomrule
 \end{tabular}}
\label{tab:auto_result}
\end{table*}

\subsection{Dataset}
We conduct our experiments on the \textsc{EmpatheticDialogues} dataset~\cite{rashkin2019towards}. 
\textsc{EmpatheticDialogues} is a large-scale multi-turn empathetic dialogue dataset collected on the Amazon Mechanical Turk, containing about 25k one-to-one open-domain conversation.
Specifically, \citet{rashkin2019towards} pair two crowd-workers: a speaker and a listener. The speaker is asked to talk about the personal emotional feelings. The listener infers the underlying emotion through what the speaker says and responds empathetically.
The dataset provides 32 evenly distributed emotion labels. 
At training time, the emotional label of the dialogue history~(i.e., the speaker) acts as a supervised signal, while we hide the label in test time to evaluate the empathetic ability of all the models.
We treat the dialogue history as the system input and the listener's response as the target output.
Then we obtain 17,802 dialogues in the training set, 2,628 in the validation set, and 2,494 in the testing set.
The average lengths of dialogue history and response are 2.1 utterances and 13.5 tokens respectively.

\subsection{Baselines for comparison}
We compare with the state-of-the-art baselines as follows:
(1) \textbf{Transformer}~\citep{vaswani2017attention}: A Transformer-based encoder-decoder model with a copy mechanism.
(2) \textbf{EmoPrepend-1}~\citep{rashkin2019towards}: An extension of the Transformer model which incorporates an additional supervised emotion classifier.
(3) \textbf{MoEL}~\citep{LinMSXF19}: Another extension of Transformer model which softly combines the response representations from different decoders. Each decoder is optimized to focus on one type of emotion accordingly.
(4) \textbf{MIME}~\citep{MajumderHPLGGMP20}: An empathetic dialogue model considering polarity-based emotion clusters and emotional mimicry.
(5) \textbf{EmpDG}~\citep{LiCRRTC20}: A multi-resolution empathetic adversarial chatbot which exploits multi-resolution emotions and user feedback.

We also conduct ablation studies to better analyze the influence of different components in our model:
(1) \textbf{w/o ECE}: The KEMP model without emotional knowledge of the emotional context encoder.
(2) \textbf{w/o EDD}: The KEMP model without emotion-dependency mechanisms of the decoder.
Additionally, we analyze the results of incorporating pre-trained model (DialoGPT~\citep{ZhangSGCBGGLD20}) in our model.

\subsection{Implementation details}
We lowercase the characters, tokenize the sequences and retain a vocabulary with 24,647 tokens.
We use pre-trained Glove vectors~\citep{pennington2014glove} to initialize the word embedding.
All common hyperparameters are the same as the work in~\citep{LiCRRTC20}. 
The maximum introducing numbers of external concepts per dialogue and per token are set as 10 and 5, respectively. The threshold $\alpha$ used in emotional context graph construction is 0.1.
Loss weights $\gamma_1, \gamma_2, \gamma_3$ are set to 1, 1, and 0.1, respectively.
We implemented all models in PyTorch~\citep{paszke2017automatic} with a single Tesla V100 GPU, and train models using Adam optimization~\citep{kingma2014adam} with a mini-batch size of 16.
We varied the learning rate during training following~\citet{vaswani2017attention}.
Early stopping is applied when training. 
When inference, we set the maximum decoding step as 30.
The training time of KEMP is 3 hours for around 26000 iterations.

\subsection{Evaluation metrics}
\paragraph{Automatic evaluations} 
To evaluate the model at the emotional level, we adopt \textbf{Emotion Accuracy} as the agreement between the ground truth emotion labels and the predicted emotion labels.
Following previous emotion-related studies~\cite{zhou2018emotional,rashkin2019towards, song2019generating, wei2019emotion,LiCRRTC20}, we adopt \textbf{Perplexity}~\cite{SerbanSBCP15}, \textbf{Distinct-1}, and \textbf{Distinct-2}~\cite{li2015diversity} to evaluate comparisons in our experiments:
Perplexity measures the high-level general quality of the generation model.
Distinct-1 / Distinct-2 is the proportion of the distinct unigrams / bigrams in all the generated results to indicate the diversity.

\paragraph{Human evaluations}
We randomly sample 100 dialogues and their corresponding generations from our model as well as the baselines. We recruit three professional annotators from a third-party company to evaluate the responses generated by different models.
All models are evaluated in terms of following 3 metrics: \textbf{Empathy}, \textbf{Relevance} and \textbf{Fluency} \cite{LinMSXF19,MajumderHPLGGMP20,LiCRRTC20}.
Empathy measures whether the generated responses express the appropriate emotions;
Relevance evaluates whether the responses are on-topic with the dialogue history;
Fluency measures the grammatical correctness and readability of the generated responses.
Each metric is rated on five-scale, where 1, 3, and 5 indicate unacceptable, moderate, and excellent performance, respectively.

\begin{table}[!t]
\caption{Ablation study.}
\centering
\resizebox{\columnwidth}{!}{
\begin{tabular}{l |c | c | c | c}
  \toprule
  \multirow{1}{*}{\textbf{Models}} &
  \multirow{1}{*}{\textbf{Accuracy}} &
  \multirow{1}{*}{\textbf{Perplexity}} &
  \multirow{1}{*}{\textbf{Distinct-1}} &
  \multirow{1}{*}{\textbf{Distinct-2}} 
  \\
  \midrule
  KEMP &\textbf{39.31} & 36.89 & \textbf{0.55} & \textbf{2.29} \\
  w/o ECE & 38.80 & 36.42 & 0.52 & 2.09 \\
  w/o EDD  & 35.41 & \textbf{36.14}  & 0.41 & 2.04 \\
  \bottomrule
\end{tabular}
}
\label{tab:abla_tests}
\end{table}

\begin{table}[!t]
\small
\caption{Result of human A/B test.} 
\centering
\resizebox{\columnwidth}{!}{
\begin{tabular}{l|c|c|c}
  \toprule
  \multirow{1}{*}{\textbf{Models}} &
   \multirow{1}{*}{\textbf{Win}} &
  \multirow{1}{*}{\textbf{Loss}} &
  \multirow{1}{*}{\textbf{Tie}} 
  \\
  \midrule
  KEMP vs Transformer & 43.8\% & 17.5\% & 38.7\% \\
  KEMP vs EmoP & 40.6\% & 18.5\% & 40.9\% \\
  KEMP vs MoEL  & 38.3\% & 18.0\% & 43.7\% \\
  KEMP vs MIME & 36.6\% & 20.6\% & 42.8\% \\
  KEMP vs EmpDG & 35.5\% & 21.3\% & 43.2\% \\
  \bottomrule
\end{tabular}
}
\label{tab:ab_result}
\end{table}

\section{Results and analysis}

\subsection{Automatic evaluation results} 
In Table~\ref{tab:auto_result}, we observe that our model KEMP outperforms strong baselines MIME and EmpDG by a large margin in terms of all automatic metrics.
The noticeable improvement indicates the effectiveness of our knowledge-enhanced model in empathetic expression and response diversity.
EmpPrepend-1 and MoEL have similar performance, as both of them only use the dialogue history to infer emotional states and generate responses. 
Without emotion modelling, Transformer only generates fluent responses based on semantic mapping, but fail to express diverse responses.

We also perform an ablation study for better understanding the contributions of the main parts of our model. 
As shown in Table~\ref{tab:abla_tests}, after we replace emotional context encoder with vanilla transformer encoder~(w/o ECE model), both the emotion accuracy and distinct performance become obviously worse, indicating that injecting external knowledge is consistently critical for emotion understanding and response generation.
We also investigate the effect of replacing emotion-dependency decoder with vanilla transformer decoder~(i.e., w/o EDD model). We notice that the scores decrease dramatically on most metrics, which demonstrates the effectiveness of modelling emotional dependencies.  

\subsection{Human evaluation results}
Table~\ref{tab:auto_result} illustrates that KEMP obtains the best performance on both Empathy and Relevance scores.
This suggests that the knowledge-enriched emotional context encoder and emotion-dependency decoder to capture implicit emotions, improve the topic consistency, and elicits a more appropriate response.
We see there is no obvious difference among models in terms of Fluency. 
We deduce it's because the generated responses by Transformer are already fluent and grammatical.
Additionally, we carried out pairwise response comparison to directly compare the dialogue quality gains in Table~\ref{tab:ab_result}.
The results confirm that the responses from KEMP are more preferred by human judges.

\begin{figure}[!t]
 \centering
 \includegraphics[width=0.4\textwidth,height=3.8cm]{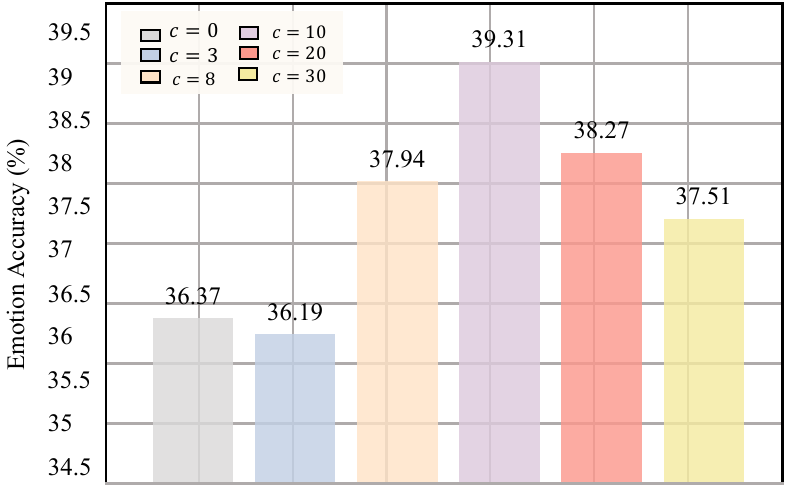}
 \caption{Emotion accuracy with respect to the maximum number of external concepts injection ($c$).}
 \label{fig:know_curve}
\end{figure}

\subsection{External knowledge analysis} 
To further investigate the impact of the different introduced number of external knowledge, we train KEMP with different numbers of concepts in terms of Accuracy. The result is shown in Figure~\ref{fig:know_curve}.
With increasing the number of concepts, the performance is rising.
However, if we introduce too many concepts, the accuracy no longer increases or even decreases. 
Therefore, external knowledge is more suitable to be the auxiliary information to perceive the emotional states in the dialogue history.

\begin{table}[!t]
\small
\caption{The visualization of the cross-attention weights in EmpDG and KEMP.}
\centering
\begin{tabularx}{0.49\textwidth}{l|l}
\toprule
{\bf History}&\colorbox[rgb]{1.0,0.8941,0.7686}{It}\colorbox[rgb]{1.0,0.8941,0.7686}{inspires}\colorbox[rgb]{0.9921,0.9608,0.9020}{me}\colorbox[rgb]{1.0,0.8078,0.6235}{to} \colorbox[rgb]{1.0,0.8941,0.7686}{try}\colorbox[rgb]{0.9921,0.9608,0.9020}{and} \colorbox[rgb]{1.0,0.8941,0.7686}{do}\\
&\colorbox[rgb]{0.9921,0.9608,0.9020}{something}
\colorbox[rgb]{0.9921,0.9608,0.9020}{to} \colorbox[rgb]{0.9921,0.9608,0.9020}{keep}\colorbox[rgb]{1.0,0.8941,0.7686}{healthy}\colorbox[rgb]{0.9921,0.9608,0.9020}{every}\colorbox[rgb]{0.9921,0.9608,0.9020}{day}. \\
EmpDG  &  I am sorry to hear. What kind of health is it? \\
\midrule
{\bf History}&\colorbox[rgb]{0.9921,0.9608,0.9020}{It}\colorbox[rgb]{1.0,0.9373,0.8353}{inspires}\colorbox[rgb]{0.9921,0.9608,0.9020}{me}\colorbox[rgb]{0.9921,0.9608,0.9020}{to}\colorbox[rgb]{1.0,0.8941,0.7686}{try}\colorbox[rgb]{0.9921,0.9608,0.9020}{and}\colorbox[rgb]{0.9921,0.9608,0.9020}{do}\\
&\colorbox[rgb]{0.9921,0.9608,0.9020}{something}\colorbox[rgb]{0.9921,0.9608,0.9020}{to}\colorbox[rgb]{1.0,0.8078,0.6235}{keep}\colorbox[rgb]{1.0,0.8078,0.6235}{healthy}\colorbox[rgb]{0.9921,0.9608,0.9020}{every}\colorbox[rgb]{0.9921,0.9608,0.9020}{day}. \\
{\bf Knowledge}& \colorbox[rgb]{1.0,0.9373,0.8353}{effort},\colorbox[rgb]{1.0,0.8941,0.7686}{fight},\colorbox[rgb]{0.9921,0.9608,0.9020}{good},\colorbox[rgb]{0.9921,0.9608,0.9020}{life},\colorbox[rgb]{0.9921,0.9608,0.9020}{raise},\colorbox[rgb]{1.0,0.8941,0.7686}{grow},\\
&\colorbox[rgb]{0.9921,0.9608,0.9020}{protect},\colorbox[rgb]{0.9921,0.9608,0.9020}{health}\\
KEMP &  I can not wait to \underline{try} to get a little \underline{makes} me\\ &\underline{feel} \underline{better}.\\
\bottomrule
\end{tabularx}
\label{tab:attn_dist}
\end{table}

\begin{table}[!t]
  \caption{Results on the pre-trained models.}
  \centering
  \resizebox{\columnwidth}{!}{
  \begin{tabular}{l |c | c | c | c}
    \toprule
     \multirow{1}{*}{\textbf{Models}} &
    \multirow{1}{*}{\textbf{Accuracy}} &
    \multirow{1}{*}{\textbf{Perplexity}} &
    \multirow{1}{*}{\textbf{Distinct-1}} &
    \multirow{1}{*}{\textbf{Distinct-2}} 
    \\
    \midrule
    KEMP-big & 45.91 & - & 2.22 & 4.93  \\
    DialoGPT & - & 15.57 & 1.57 & 4.18  \\
    KEMP-DialoGPT & \textbf{46.43} & \textbf{15.21} & \textbf{2.79} & \textbf{4.24} \\
    \bottomrule
  \end{tabular}
  }
  \label{tab:plm}
  \end{table}

\subsection{Emotion-dependency analysis} 
Table~\ref{tab:attn_dist} shows an example illustrating the cross-attention weights of the dialogue context.
Baseline EmpDG puts the major attention on general words, which leads to a context-inconsistent and emotion-inappropriate response.
In comparison, the KEMP model puts the highest attention probability on the words containing informative meaning, e.g., ``\textit{fight}'' and ``\textit{grow}'' in external knowledge and ``\text{keep}'' and ``\textit{healthy}'' in dialogue history.
We can conclude that the proposed emotion-dependency mechanism in the decoder can teach the model to generate responses from meaningful and emotional words.

\begin{table}[!t]
\small
\caption{Generated responses from KEMP and baseline models in two different speaker's emotion states. Tokens in \underline{underline} represent knowledge-related words.
}
\renewcommand\arraystretch{0.8}
\centering
\begin{tabular}{l|l}
\toprule
{\bf Emotion}&{\bf Terrified}\\
{\bf History}&\multicolumn{1}{p{0.75\columnwidth}}{$X_1$: Do you know how crazy it is to skydive?}\\
{}&\multicolumn{1}{p{0.75\columnwidth}}{$X_2$: I have a fear of falling from high places.}\\
{}&\multicolumn{1}{p{0.75\columnwidth}}{$X_3$: It gave me the biggest rush that is for sure.}\\
\midrule
\textbf{Gold} & \textbf{I think I would pass out from fear lol.} \\
Transformer & I am sure it was.\\
EmoP  & I am sure it was a lot of fun.  \\
MoEL & I am sure you will do it!  \\
MIME & That is a good thing to do.  \\
EmpDG & I think it is an amazing feeling. \\
\dashrule & \dashrule\\
{Know.}&\multicolumn{1}{p{0.75\columnwidth}}{danger, pain, travel, \underline{scared}, excited, \underline{furious}}\\
KEMP & I would have been so \underline{scared}. \\
\midrule
{\bf Emotion}&{\bf Surprised}\\
{\bf History}& \multicolumn{1}{p{0.75\columnwidth}}{$X_1$: \ I realized yesterday I was pregnant, I was in shock!}\\
\midrule
\textbf{Gold} & \multicolumn{1}{p{0.7\columnwidth}}{\textbf{Woah, that is huge news. How are you handling the news.}} \\
Transformer & Oh no! I am so sorry to hear that.\\
EmoP  & Oh no! Did you get a job?  \\
MoEL & That is so awesome! I am glad to hear that.  \\
MIME & That is great.  \\
EmpDG & Oh wow! what happened? \\  
\dashrule & \dashrule\\
{Know.}&{experience, \underline{surprise}, mother, pain, \underline{feeling}}\\
KEMP & Oh wow! That is a \underline{great} \underline{feeling}. \\
\bottomrule
\end{tabular}
\label{tab:cases}
\end{table}

\subsection{Effectiveness of pre-trained model}
As show in Table~\ref{tab:plm}, we also explore if we can improve the performance by integrating KEMP with the pre-trained model on dialogues, i.e, DialoGPT~\cite{ZhangSGCBGGLD20}.  
KEMP-big is the KEMP with the same transformer hyperparameters setups as the DialoGPT.
KEMP-DialoGPT incorporates the graph-attention layer of emotional context encoder and the cross-attention layer of emotion-dependency decoder into the DialoGPT.
We can find that pre-trained models is effective in the empathetic dialogue genration because of the huge amount of pre-trained dialogue datasets.
More important,
we see that KEMP-DialoGPT outperforms DialoGPT, which concludes injecting emotional knowledge is able to improve the generation performance.
\vspace{-2mm}

\subsection{Case study}
Cases from KEMP and baseline models are listed in Table~\ref{tab:cases}.
In the first case, KEMP generates informative responses with a proper negative emotion by replying with ``\textit{scared}''. However, without emotional knowledge, all baselines fail to recognize the negative emotion. 
In the second case, KEMP model generates the most context-consistent response, which contains context-related word~(``\textit{feeling}'') and emotion-rated word~(``\textit{Oh wow}'').
Both the two cases show that the KEMP can balance the performances between content and emotion.

\section{Conclusion and outlook}
In this work, we have proposed a knowledge-aware empathetic dialogue generation model, KEMP, to enhance the emotion perception and dependencies abilities of empathetic dialogue system with bunches of emotion-related concepts.
Experimental results show that KEMP outperforms state-of-the-art methods in terms of both automatic and human evaluations.
Besides, we verify the effectiveness of the emotional context graph, emotional context encoder, and the emotion-dependency decoder in KEMP. 

KEMP adopts heuristic rules to construct emotional context graph, which is not flexible to adapt different knowledge resources.
As for future work,  we plan to address this issue by integrating with knowledge reasoning models to automatically construct emotional context graph.

\newpage
\section{Acknowledgments}
We want to thank our anonymous reviewers for their feedback.
This work was supported by the National Key R\&D Program of China with grant No. 2020YFB1406704, the Natural Science Foundation of China (62106105, 62102234, 62072279, 61902219, 61972234), the Key Scientific and Technological Innovation Program of Shandong Province (2019JZZY010129), the Natural Science Foundation of Shandong Province (ZR2021QF129), the Fundamental Research Funds of Shandong University.
\bibliography{aaai22}

\end{document}